\documentclass[final]{cvpr}

\usepackage{times}
\usepackage{epsfig}
\usepackage{graphicx}
\usepackage{amsmath}
\usepackage{amssymb}
\usepackage{kotex}
\usepackage{booktabs}
\usepackage{multirow}
\usepackage{dsfont}
\usepackage[dvipsnames]{xcolor}
\usepackage{subcaption}
\usepackage{pifont}
\usepackage{floatrow}
\usepackage{nicefrac}
\usepackage{algorithm,algorithmicx,algpseudocode}

\newfloatcommand{capbtabbox}{table}[][\FBwidth]
\usepackage[pagebackref=true,breaklinks=true,colorlinks,bookmarks=false]{hyperref}

\def\cvprPaperID{4535} %
\def\confYear{CVPR 2021}

\newcommand\ours{{{\mbox{ReLabel}}}\xspace}
\newcommand\oursb{{\textbf{\mbox{ReLabel}}}\xspace}
\newcommand\ourframework{{{LabelPooling}}\xspace}
\newcommand\ourframeworkb{{\textbf{{LabelPooling}}}\xspace}

\definecolor{darkergreen}{RGB}{21, 152, 56}
\definecolor{red2}{RGB}{252, 54, 65}

\definecolor{airforceblue}{rgb}{0.36, 0.54, 0.66}

\begin{document}

\title{
Re-labeling ImageNet: \\ from Single to Multi-Labels, from Global to Localized Labels}

\author{
Sangdoo Yun~~~~Seong Joon Oh~~~~Byeongho Heo~~~~Dongyoon Han~~~~Junsuk Choe~~~~Sanghyuk Chun\\
~~\\
NAVER AI Lab\\
}

\maketitle

\begin{abstract}
ImageNet has been the most popular image classification benchmark, but it is also the one with a significant level of label noise. Recent studies have shown that many samples contain multiple classes, despite being assumed to be a single-label benchmark. They have thus proposed to turn ImageNet evaluation into a multi-label task, with exhaustive multi-label annotations per image. However, they have not fixed the training set, presumably because of a formidable annotation cost. 
We argue that the mismatch between single-label annotations and effectively multi-label images is equally, if not more, problematic in the training setup, where random crops are applied. With the single-label annotations, a random crop of an image may contain an entirely different object from the ground truth, introducing noisy or even incorrect supervision during training. 
We thus re-label the ImageNet training set with multi-labels. We address the annotation cost barrier by letting a strong image classifier, trained on an extra source of data, generate the multi-labels. We utilize the pixel-wise multi-label predictions before the final pooling layer, in order to exploit the additional location-specific supervision signals. Training on the re-labeled samples results in improved model performances across the board. 
ResNet-50 attains the top-1 accuracy of \textbf{78.9\%} on ImageNet with our localized multi-labels, which can be further boosted to \textbf{80.2\%} with the CutMix regularization. We show that the models trained with localized multi-labels also outperforms the baselines on transfer learning to object detection and instance segmentation tasks, and various robustness benchmarks. The re-labeled ImageNet training set, pre-trained weights, and the source code are available at 
{\url{https://github.com/naver-ai/relabel_imagenet}}.
\end{abstract}

\section{Introduction}
The ImageNet dataset~\cite{ImageNet} has been at the center of modern advances in computer vision.
Since the introduction of ImageNet,
image recognition models based on convolutional neural networks have made quantum jumps in performances~\cite{alexnet, VGG, resnet}.
Improving the model performance on ImageNet is seen as a litmus test for the general applicability of the model and the transfer learning performances on downstream tasks~\cite{kornblith2019better,yun2019cutmix}.

\begin{figure}[t]
\small
\centering
\includegraphics[width=1.0\linewidth]{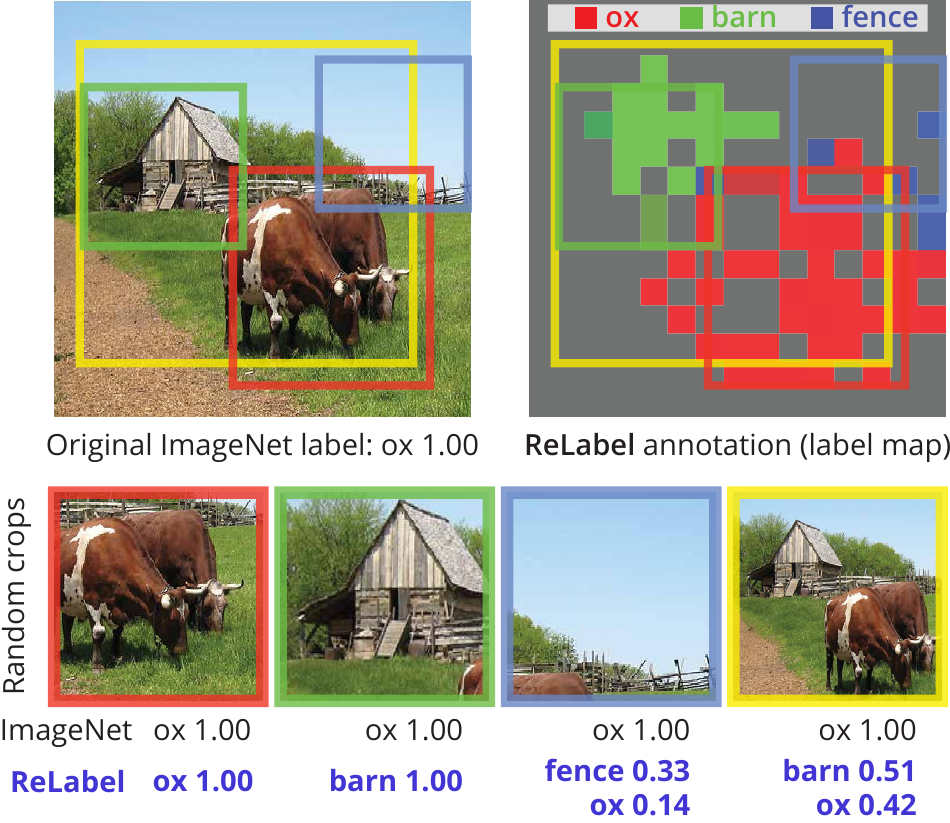}
\caption{\textbf{Re-labeling ImageNet training data.} 
Original ImageNet annotation is a single label (``ox''), whereas the image contains multiple ImageNet categories (``ox'', ``barn'', and ``fence''). 
Random crops of an image may contain an entirely different object category from the global annotation. 
Our method (\oursb) generates location-wise multi-labels, resulting in cleaner supervision per random crop.
}
\label{fig:teaser}
\vspace{-0.2cm}
\end{figure}

ImageNet, however, turns out to be noisier than one would expect. Recent studies~\cite{stock2018convnets,tsipras2020imagenet_madry,beyer2020are_we_done,shanker2020machine_accuracy} have shed light on an overlooked problem with ImageNet that a significant portion of the dataset is composed of images with multiple possible labels. This contradicts the underlying assumption that there is only a single object class per image: the evaluation metrics penalize any prediction beyond the single ground-truth class. 
Thus, researchers have refined the ImageNet validation samples with multi-labeling policy using human annotators~\cite{beyer2020are_we_done,shanker2020machine_accuracy}, and proposed new multi-label evaluation metrics.
Under these new evaluation schemes, recent state-of-the-art models~\cite{xie2020noisy_student,touvron2019fixing} that seem to have surpassed the human level of recognition have been found to fall short of the human performance level.

The mismatch between the multiplicity of object classes per image and the assignment of single labels results in problems not only for evaluation, but also for training: the supervision becomes noisy.
The widespread adoption of \textit{random crop} augmentation~\cite{GoogleNet} aggravates the problem. 
A random crop of an image may contain an entirely different object from the original single label, introducing potentially wrong supervision signals during training, as in Figure~\ref{fig:teaser}. 

\begin{figure}[t]
\centering
\includegraphics[width=\linewidth]{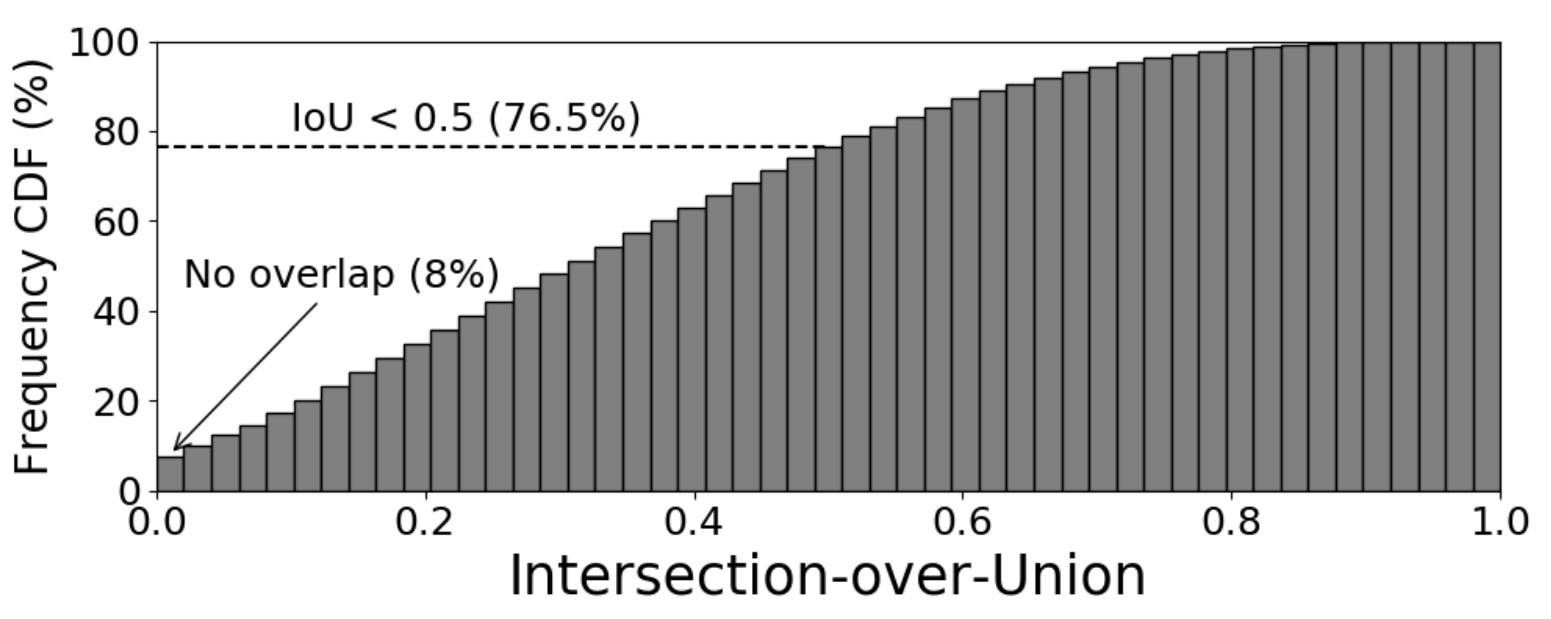}
\caption{\textbf{Cumulative distribution of Intersection-over-Union (IoU)} between the random crops and ground-truth bounding boxes. We sample 100 random crops per image on the ImageNet validation set ($50$K images).
}
\label{fig:rrc_overlap}
\vspace{-0.2cm}
\end{figure}

The random crop augmentation makes supervision noisy not only for images with multiple classes. Even for images with a single class, the random crop often contains no foreground object. It is estimated that, under the standard training setup\footnote{A random crop is sampled from $8$\% to $100\%$ of the entire image area.}, $8\%$ of the random crops have no overlap with the ground truths. Only $23.5\%$ of the random crops have the intersection-over-union (IoU) measure greater than $50\%$ with the ground truth boxes (see Figure~\ref{fig:rrc_overlap}). Training a model on ImageNet inevitably involves a lot of noisy supervision.

Ideally, for each training image, we want a human annotation telling the model (1) the full set of classes present (multi-label) and (2) where each object is located (localized label). One such format would be a dense pixel labeling $L\in \{0,1\}^{H\times W\times C}$ where $C$ is the number of classes, as done for semantic segmentation ground truths. However, it is hardly scalable to collect even just the multi-label annotations for the 1.28 million ImageNet training samples. It took more than three months for five human experts (authors of \cite{shanker2020machine_accuracy}) to label mere 2,000 images.

In this paper, we propose a re-labeling strategy, \oursb, to obtain pixel-wise labeling $L\in\mathbb{R}^{H\times W\times C}$, which are both multi-labels and localized labels, on the ImageNet training set. We use strong classifiers trained on external training data to generate those labels. The predictions before the final pooling layer have been used.
We also contribute a novel training scheme, \ourframeworkb, for training classifiers based on the dense labels. For each random crop sample, we compute the multi-label ground truth by pooling the label scores from the crop region.
\ours incurs only a one-time cost for generating the label maps per dataset, unlike \eg Knowledge Distillation~\cite{hinton2015distilling} which involves one forward pass per training iteration to generate the supervision. Our \ourframework supervision adds only a small amount of computational cost on the usual single-label cross-entropy supervision.

We present an extensive set of evaluations for various model architectures trained with \ours on multiple datasets and tasks. 
On ImageNet classification, training ResNet-50 with ImageNet \ours has achieved a top-1 accuracy of \textbf{78.9\%}, a \textbf{+1.4 pp} gain over the baseline model trained with the original labels. The accuracy of ResNet-50 reaches \textbf{80.2\%} by employing the CutMix regularization on top, a new state-of-the-art performance on ImageNet to the best of our knowledge. 
Models trained with \ours have also consistently improved accuracies on ImageNet multi-label evaluation metrics proposed by \cite{beyer2020are_we_done,shanker2020machine_accuracy}.
\ours and \ourframework result in consistent improvements for transfer learning experiments, including the object detection and instance segmentation tasks on COCO and fine-grained classifications tasks. 
We further test {\ourframework} on the multi-label classification task on COCO. 
Finally, we show that models trained with \ours are more resilient to test-time perturbations, as will be verified through experiments on several robustness benchmarks.

\section{Related Works}

We start this section by introducing prior works discussing the issues with ImageNet labels. We then discuss a few other research areas that share similarities with our approach. We describe the key differences from our approach.

\noindent\textbf{Labeling issues in ImageNet.}
ImageNet~\cite{ImageNet} has effectively served as the standard benchmark for the image classifiers: ``methods live or die by their performance on this benchmark'', as argued by Shankar \etal~\cite{shanker2020machine_accuracy}. 
The reliability of the benchmark itself has thus come to be the subject of careful research and analysis. 
As with many other datasets, ImageNet contains much label noise~\cite{van2015building,recht2019imagenetv2}.
One of the most persistent and systematic types of label error on ImageNet is the erroneous single labels~\cite{stock2018convnets,shanker2020machine_accuracy,tsipras2020imagenet_madry,beyer2020are_we_done}, referring to the cases where only one out of multiple present categories is annotated. Such errors are prevalent, as ImageNet contains many images with multiple classes. Shankar \etal~\cite{shanker2020machine_accuracy} and Beyer \etal~\cite{beyer2020are_we_done} have identified three subcategories for the erroneous single labels: (1) an image has multiple object classes, (2) there exist multiple labels that are synonymous or hierarchically including the other, and (3) inherent ambiguity in an image makes multiple labels plausible. 
Those studies have refined the validation set labels into multi-labels to establish an truthful and fair evaluation of models on effectively multi-label images.
The focus of \cite{shanker2020machine_accuracy}, however, has been only the validation, not training. \cite{beyer2020are_we_done} has introduced a clean-up scheme to remove training samples with potentially erroneous labels by validating them with predictions from a strong classifier. 
Our work focuses on the clean-up strategy for the ImageNet training labels. 
Like \cite{beyer2020are_we_done}, we utilize strong classifiers to clean up the training labels. Unlike \cite{beyer2020are_we_done}, we \emph{correct} the wrong labels, not \emph{remove}. Our labels are also given per region. 
In our experiments, our method shows improved results compared to \cite{beyer2020are_we_done}.

\begin{table}[t]
\small
\centering
\tabcolsep=0.08cm
\begin{tabular}{@{}lccc@{}}
\toprule
                            & multi-label & local label & efficient \\ \midrule
Original ImageNet training    & \textcolor{red2}{\ding{56}}   & \textcolor{red2}{\ding{56}}      & \textcolor{darkergreen}{\ding{52}}      \\
Knowledge distillation & \textcolor{darkergreen}{\ding{52}}     & \textcolor{darkergreen}{\ding{52}}      & \textcolor{red2}{\ding{56}}    \\
\ours \& \ourframework (ours) & \textcolor{darkergreen}{\ding{52}}    & \textcolor{darkergreen}{\ding{52}}     & \textcolor{darkergreen}{\ding{52}}    \\ 
\bottomrule
\end{tabular}
\caption{\textbf{What are the differences?} Comparison among the training options on ImageNet.
}
\label{tab:checkbox}
\vspace{-0.3cm}
\end{table}

\noindent\textbf{Knowledge distillation.}
Knowledge distillation (KD)~\cite{hinton2015distilling} also utilizes machine supervisions generated by the ``teacher'' network. 
Studies on KD have enriched and diversified the options for the teacher, such as feature map distillation~\cite{zagoruyko2016paying,heo2019AB,heo2019comprehensive}, relation-based distillation~\cite{park2019relational,tian2019contrastive}, ensemble distillation~\cite{shen2019meal,zhu2018knowledge}, or iterative self-distillation~\cite{furlanello2018born,yang2019snapshot,xie2020noisy_student}.
While those studies pursue stronger forms of supervision, none of them have considered a strong, state-of-the-art network as a teacher because it makes the KD supervision far heavier and impractical. With the random crop augmentation in place, every training iteration would involve a forward pass through the strong yet heavy teacher. 
Ours is similar in that the model is trained with machine supervision, but is more efficient\footnote{From a more general view of KD that utilizes teacher and student in any form, our method can be seen as a new and efficient type of KD.}. 
\ourframework supervises a network with pre-computed label maps, rather than generating the label on the fly through the teacher for every random crop during training. 
We present the key advantages of ours against KD in Table~\ref{tab:checkbox}.

\noindent\textbf{Training tricks for ImageNet.}
Data augmentation is a simple yet powerful strategy for ImageNet training.
The standard augmentation setting includes random cropping, flipping, and color jittering, as used in~\cite{pyramidnet,ghiasi2018dropblock,densenet,GoogleNet,yun2019cutmix,touvron2019fixing}. 
In particular, the random crop augmentation, which crops random coordinates in an image and resize to a fixed size, is indispensable for a reasonable performance on ImageNet. 
Our work considers localized labels that make the supervision provided for each random crop region more sensible.
There are additional training tricks for training classifiers~\cite{zhang2017mixup,yun2019cutmix,ghiasi2018dropblock,stochasticdepth,devries2017cutout} that are orthogonal to our re-labeled training data.
We show that those tricks can be combined with our re-labeling for improved performances.

\section{Method}

We propose a re-labeling strategy \oursb to obtain pixel-level ground truth labels on the ImageNet training set. The label maps have two characteristics: (1) multi-class labels and (2) localized labels. The labels maps are obtained from a machine annotator: a strong image classifier trained on an extra data.
We describe how to obtain the label maps and present a novel training framework, \ourframeworkb, to train image classifier using such localized multi-labels.

\subsection{Re-labeling ImageNet}

We obtain dense ground truth labels from a \textit{machine annotator}, a state-of-the-art classifier that has been pre-trained on a super-ImageNet scale (\eg JFT-300M~\cite{sun2017revisiting} or InstagramNet-1B~\cite{InstagramNet}) and fine-tuned on ImageNet to predict ImageNet classes.
Predictions from such a model are arguably close to human predictions~\cite{beyer2020are_we_done}. %
Since training the machine annotators requires an access to proprietary training data~\cite{sun2017revisiting,InstagramNet} and hundreds of GPU or TPU days, we have adopted the open-source trained weights as the machine annotators.
We show the comparison of different available machine annotators later in Section~\ref{subsec:discussion}.

We remark that while the machine annotators are trained with single-label supervision on ImageNet, they still tend to make multi-label predictions for images with multiple categories. As an illustration, consider an image $x$ with two correct categories $0$ and $1$. Assume that the model is fed with both $(x,y=0)$ and $(x,y=1)$ equal number of times during training, with those noisy labels. Then, the cross-entropy loss is given by $-\frac{1}{2}(\sum_k y^0_k\log p_k(x) + \sum_k y^1_k\log p_k(x)) = -\sum_k \frac{y^0_k+y^1_k}{2}\log p_k(x)$ where $y^c$ is the one-hot vector with $1$ at index $c$ and $p(x)$ is the prediction vector for $x$. Note that the minimal value for the function $-\sum_k q_k \log p_k$ with respect to $p$ is taken at $p=q$. Thus, in this example, the model minimizes the loss by predicting $p(x)=(\frac{1}{2},\frac{1}{2})$.
Thus, if there exist much label noise in the dataset, a model trained with the single-label cross-entropy loss tends to predict multi-label outputs.

As an additional benefit of obtaining labels from a classifier, we consider extracting the location-specific labels. 
We remove the \emph{global average pooling} layer of the classifier and turn the following linear layer into a $1\times 1$ convolutional layer, thereby turning the classifier into a fully-convolutional network~\cite{zhou2016CAM,FCN}. 
The output of the model then becomes $f(x)\in\mathbb{R}^{W\times H\times C}$. 
We use the output $f(x)$ as our \textit{label map} annotations $L\in\mathbb{R}^{W\times H\times C}$.
We present the detailed procedure to obtain label maps in Appendix~B. %

\begin{figure}[t]
\centering
\includegraphics[width=\linewidth]{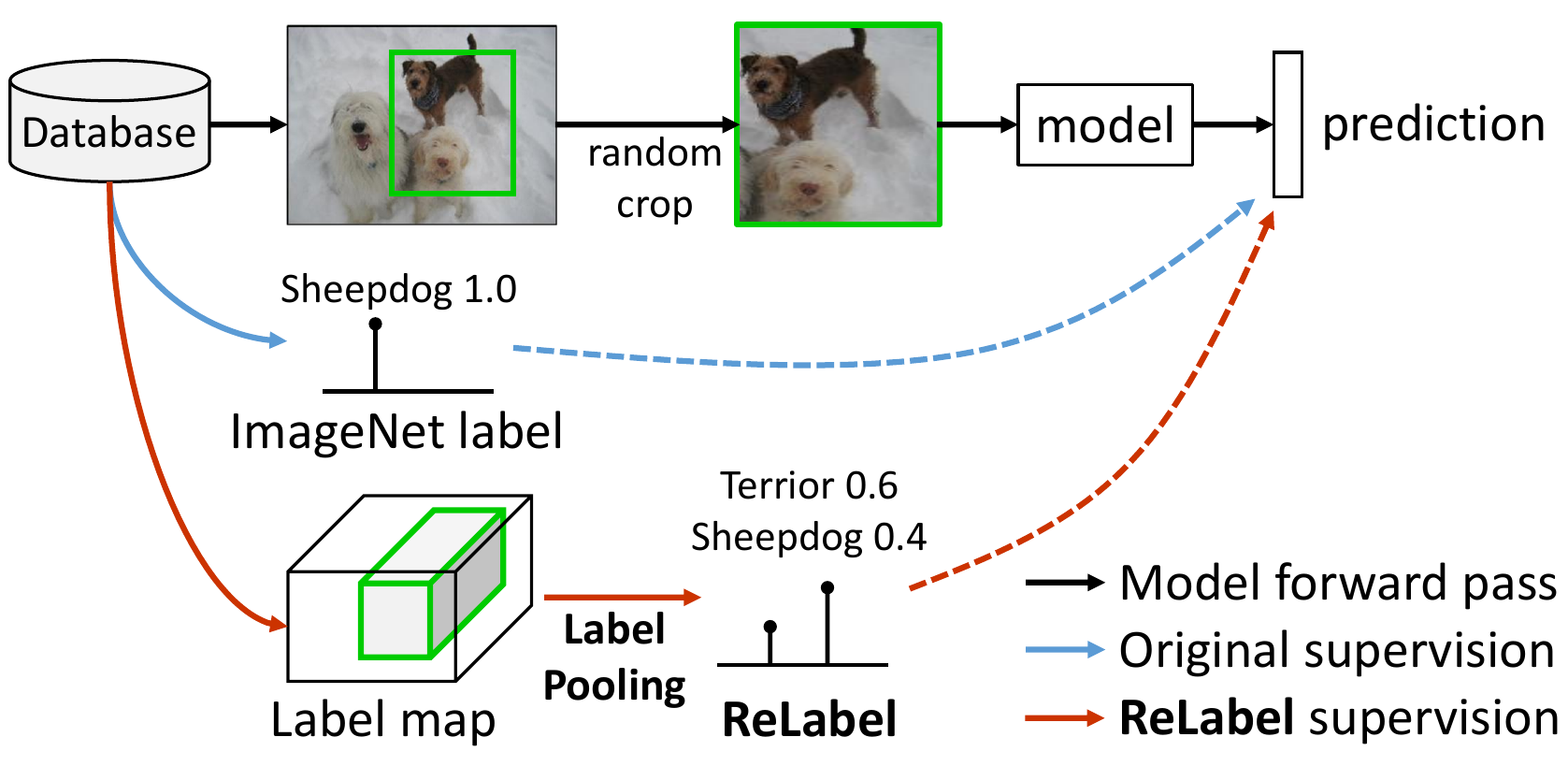}
\caption{
\textbf{Illustration of  \ourframework.}
Original ImageNet supervision is single-label (``Sheepdog''). \ourframework trains the model with \ours, localized multi-labels, (``Sheepdog'' and ``Terrior'') based on the crop region.
}
\label{fig:framework}
\vspace{-0.2cm}
\end{figure}

\subsection{Training a Classifier with Dense Multi-labels}
Having obtained the dense multi-labels $L\in\mathbb{R}^{W\times H\times C}$ as above, how do we train a classifier with them? 
For this, we propose a novel training scheme, \ourframework, that takes the localized ground truths into account. 
We show the difference between \ourframework and the original ImageNet training in Figure~\ref{fig:framework}.
In a standard ImageNet training setup, the supervision for the randomly crop is given by the single label ground truth given per image. 
On the other hand, \ourframework loads a pre-computed label map and conducts a regional pooling operation on the label map corresponding to the coordinates of the random crop.
We adopt the \texttt{RoIAlign}~\cite{he2017mask} regional pooling approach.
Global average pooling and \texttt{softmax} operations are performed on the pooled prediction maps to get a multi-label ground-truth vector in $[0,1]^C$ with which the model is trained. We use the cross-entropy loss. 
Code-level implementation of our training scheme is presented in Appendix~A.

\subsection{Discussion}
\label{subsec:discussion}

So far we have introduced our labeling strategy and the supervision scheme using the label maps. We study the space and time consumption for our approach and examine design choices.

\noindent\textbf{Space consumption.}
We utilize EfficientNet-L2~\cite{xie2020noisy_student} as the machine annotator whose input resolution is $475\times475$ and the resulting label map dimension is $L\in\mathbb{R}^{15\times 15\times 1000}$. 
Saving the entire label maps for all classes will require more than 1 TB of storage: $(1.28\times10^6) \,\text{images}\,\times(15\times15\times1000)\,\nicefrac{\text{dim}}{\text{image}}\, \times \,4\nicefrac{\text{bytes}}{\text{dim}} \approx 1.0 \,\text{TB}$.
Fortunately, for each image, pixel-wise predictions beyond a few top-$k$ classes are essentially zero. Hence, we save the storage space by storing only the top-5 predictions per image, resulting in {10} GB of label map data. This corresponds to only 10\% additional space on top of the original ImageNet data.

\noindent\textbf{Time consumption.}
\ours requires a one-time cost for forward passing the ImageNet training images through the machine annotator. This procedure takes about 10 GPU-hours, which is only {3.3\%} of the entire train time for ResNet-50 (328 GPU-hours\footnote{300 epochs on four NVIDIA V100 GPUs.}). 
For each training iteration, \ourframework performs the label map loading and regional pooling operations on top of the standard ImageNet supervision, which leads to only {0.5\%} additional training time. 
Note that \ours is much more computationally efficient than knowledge distillation which requires a forward pass through the teacher at every iteration. For example, KD with EfficientNet-B7 teacher takes more than four times the original training time.

\begin{figure}[t]
\centering
\includegraphics[width=0.9\linewidth]{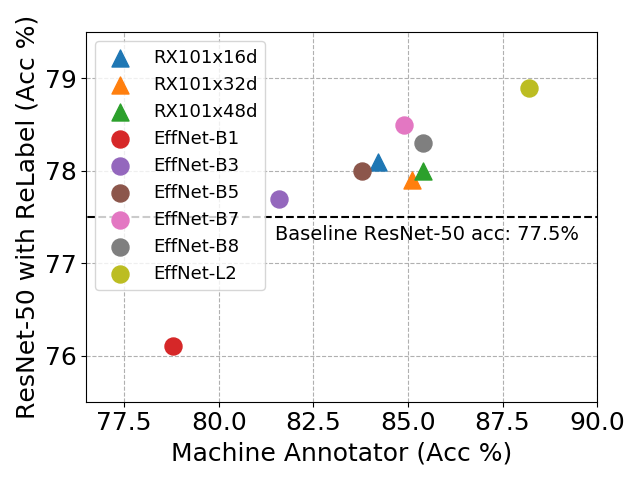}
\caption{\textbf{Machine annotators.} We plot the top-1 accuracy of ResNet-50 trained with \ours, where \ours is generated by various machine annotators.
}
\label{fig:strong_cls_ablation}
\vspace{-0.25cm}
\end{figure}

\noindent\textbf{Which machine annotator should we select?}
Ideally, we want the machine annotator to provide precise labels on training images. For this we consider \ours generated by a few state-of-the-art classifiers EfficientNet-\{B1,B3,B5,B7,B8\}~\cite{efficientnet}, EfficientNet-L2~\cite{xie2020noisy_student} trained with JFT-300M~\cite{sun2017revisiting}, and ResNeXT-101\_32x\{32d,48d\}~\cite{yalniz2019billion} trained with  InstagramNet-1B~\cite{InstagramNet}.
We train ResNet-50 with the above label maps from diverse classifiers.
Note that ResNet-50 achieves the top-1 validation accuracy of $77.5\%$ when trained on vanilla single labels.
We show the results in Figure~\ref{fig:strong_cls_ablation}.
The performance of the target model overall follows the performance of the machine annotator. 
When the machine supervision is not sufficiently strong (\eg, EfficientNet-B1), the trained model shows a severe performance drop {(76.1\%)}. 
We choose EfficientNet-L2 as the machine annotator that has led to the best performance for ResNet-50 ($78.9\%$) in the rest of the experiments.

\begin{table}
\centering
\tabcolsep=0.12cm
\begin{tabular}{@{}lc@{}}
\toprule
Variants     & ImageNet top-1 (\%) \\ \midrule
\ours (localized mutli-labels) & 78.9 \\ \midrule         
Localized single labels    & 78.4 (-0.5) \\
Global multi-labels  &  78.5 (-0.4) \\
Global single labels &  77.5 (-1.4) \\ \midrule
Original ImageNet labels &  77.5 (-1.4) \\
\bottomrule
\end{tabular}
\caption{\textbf{Factor analysis of \ours.} Results when either or both of the multi-labelness and localizability properties are removed from \ours.
}
\label{tab:factor_analysis}
\vspace{-0.2cm}
\end{table}

\noindent\textbf{Factor analysis of \ours.}
\ours is both multi-label and pixel-wise.
To examine the necessity of the two properties, we conduct an experiment by ablating each of them.
We consider the \textit{localized single labels} by taking \texttt{argmax} operation instead of \texttt{softmax} after the \texttt{RoIAlign} regional pooling, resulting in $L_{\text{loc,single}}\in \{0,1\}^{C}$.
For \textit{global multi-labels}, we take the global average pooling, instead of the \texttt{RoIAlign}, over the label map, resulting in the label $L_{\text{glob,multi}}\in [0,1]^{C}$. Finally, by first performing the global average pooling and then performing argmax, we obtain the \textit{global single-labels}, $L_{\text{glob,single}}\in \{0,1\}^{C}$. Note that $L_{\text{glob,single}}\in \{0,1\}^{C}$ labels have the same format as the original ImageNet labels, but are machine-generated.

The results for those four variants are in Table~\ref{tab:factor_analysis}.
We observe that from the \ours performance of $78.9\%$, the removal of multi-labels and localized labels results in -0.5 pp and -0.4 pp drops, respectively. When both are missing, there is a significant -1.4 pp drop. We thus argue that both ingredients are indispensable for a good performance. Note also that the global, single labels generated by a machine do not bring about any gain compared to the original ImageNet labels. This further signifies the importance of the aforementioned properties to benefit maximally from the machine annotations.

\begin{figure}[t]
\vspace{-0.25cm}
\centering
\includegraphics[width=\linewidth]{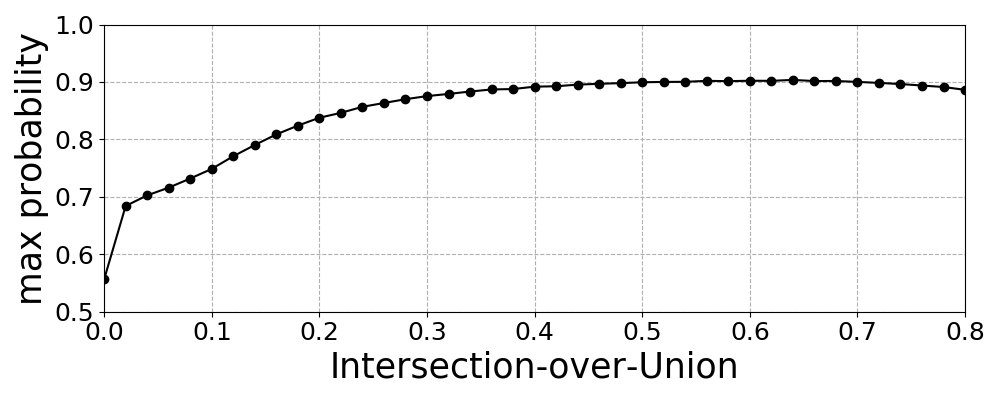}
\caption{
\textbf{\ours confidence versus GT overlap.} We plot the relationship between the confidence level for \ours pooled from the crop regions and the their overlap (IoU) with the ground truth boxes. 
}
\label{fig:rrc_score_iou}
\vspace{-0.25cm}
\end{figure}

\noindent\textbf{Confidence of \ours supervision.}
We study the confidence of \ours supervisions at different simulated levels of overlap between the random crop and the ground-truth bounding box. 
We draw $5$M random crop samples as done for Figure~\ref{fig:rrc_overlap}.
{We measure the confidence for the \ours's supervision in terms of the maximum class probability of the pooled label (\ie, $\text{confidence}=\max_c L(c)$ where $L\in[0,1]^C$).}
The results are shown in Figure~\ref{fig:rrc_score_iou}.
The averaged degree of supervision of \ours overall follows the degree of object existence, in particular, with small overlaps with object region (IoU $<0.4$).
For example, when IoU is zero (\ie, random crops are outside the object region), the label confidence is below $0.6$, providing some uncertainty signals for the trained model.

\section{Experiments}
\label{sec:experiments}

We present various experiments where we apply our labeling and training schemes for localized multi-label training.
We first show the effectiveness of \ours on ImageNet classification with various network architectures and evaluation metrics, including the recently proposed multi-label evaluation metrics and robustness benchmarks (Section~\ref{sec:exp_imagenet_cls}).
Next, we show the transfer-learning performances for models trained with \ours when they are fine-tuned for object detection, instance segmentation, and fine-grained classification tasks (Section~\ref{sec:exp_transfer}). 
We show that \ours improves the performances also for models on COCO multi-label classification tasks (Section~\ref{sec:exp_multilabel_cls}).
The re-labeled ImageNet training set, pre-trained weights, and the source code are availalble at \url{https://github.com/naver-ai/relabel_imagenet}.

\subsection{ImageNet Classification}
\label{sec:exp_imagenet_cls}

\begin{table*}
\centering
\tabcolsep=0.1cm
\begin{tabular}{@{}llcccc@{}}
\toprule
& & ImageNet &  ImageNetV2~\cite{recht2019imagenetv2} & ReaL~\cite{beyer2020are_we_done} &  Shankar \etal~\cite{shanker2020machine_accuracy}\\
Network\hspace{2em} & Supervision & {\small single-label} &  {\small single-label} & {\small multi-label} &  {\small multi-label}\\
            
\midrule
ResNet-50 & Original  & 77.5 & 79.0 & 83.6 & 85.3 \\
ResNet-50 & Label smoothing ($\epsilon$=0.1)~\cite{szegedy2016rethinking_labelsm} & 78.0 & 79.5 & 84.0 & 84.7 \\
ResNet-50 & Label cleaning~\cite{beyer2020are_we_done}  & 78.1 & 79.1 & 83.6 & 85.2 \\ 
ResNet-50 & \oursb   & \textbf{78.9} & \textbf{80.5} & \textbf{85.0} & \textbf{86.1}  \\ 
\bottomrule
\end{tabular}
\caption{\textbf{ImageNet classification.} 
Results with different types of supervision. We report performances on the single-label benchmarks (ImageNet validation set and ImageNetV2~\cite{recht2019imagenetv2}) and multi-label benchmarks (ReaL~\cite{beyer2020are_we_done} and Shankar \etal~\cite{shanker2020machine_accuracy}). 
}
\label{table:imagenet_comparison}
\vspace{-0.2cm}
\end{table*}

We evaluate \ours strategy on the ImageNet-1K~\cite{ImageNet} containing 1.28 million training images and 50,000 validation images of 1,000 object categories. 
We use standard data augmentation such as random cropping, flipping, color jittering, as in~\cite{pyramidnet,ghiasi2018dropblock,densenet,GoogleNet,yun2019cutmix,touvron2019fixing} for all the models considered.
We have trained the models with SGD for $300$ epochs with the initial learning rate $0.1$ and the cosine learning rate scheduling without restarts~\cite{loshchilov2016sgdr}. The batch size and weight decay are set to $1,024$ and $0.0001$, respectively.

\begin{table}
\vspace{-0.2cm}
\centering
\tabcolsep=0.13cm
\begin{tabular}{@{}lrrcccc@{}}
\toprule
   && \multicolumn{2}{c}{Resources}    && \multicolumn{2}{c}{Supervision}  \\ 
Architecture       && Params & Flops    && Vanilla & \ours \\ \midrule
ResNet-18 && 11.7M& 1.8B && 71.7 & 72.5 (+0.8)\\
ResNet-50 && 25.6M & 3.8B && 77.5 & 78.9 (+1.4)\\
ResNet-101 && 44.7M& 7.6B && 78.1 & 80.7 (+2.6) \\
\midrule
EfficientNet-B0 && 5.3M & 0.4B  && 77.4 & 78.0 (+0.6) \\
EfficientNet-B1 && 7.8M & 0.7B  && 79.2 & 80.3 (+1.1) \\
EfficientNet-B2 && 9.2M & 1.0B  && 80.3 & 81.0 (+0.7) \\
EfficientNet-B3 && 12.2M & 1.8B && 81.7 & 82.5 (+0.8) \\
\midrule
ReXNet ($\times$1.0) && 4.8M & 0.4B  && 77.9 & 78.4 (+0.5) \\
\bottomrule
\end{tabular}
\caption{\textbf{\ours on multiple architectures.} Validation top-1 results when supervised with the original labels (Vanilla) and \ours.}
\label{table:imagenet_various_arch}
\vspace{-0.2cm}
\end{table}

\noindent\textbf{Comparison against other label manipulations.}
We compare \ours against prior methods that directly adjust the ImageNet labels. Label smoothing~\cite{szegedy2016rethinking_labelsm} assigns a slightly weaker weight on the foreground class ($1-\epsilon$) and distributes the remaining weight $\epsilon$ uniformly across background classes.
Label cleaning by Beyer \etal~\cite{beyer2020are_we_done} prunes out all training samples where the ground truth annotation does not agree with the prediction of a strong teacher classifier, namely BiT-L~\cite{kolesnikov2019big}. For this, we use the list of clean sampled provided by the authors~\cite{beyer2020are_we_done} with our own training setting.
We conducted the above label manipulation methods and \ours on ResNet-50. 
Results are given in Table~\ref{table:imagenet_comparison}. 
We measure the single-label accuracies on ImageNet validation and ImageNetV2 ({\small Top-Images}~\cite{recht2019imagenetv2}\footnote{Results on ImageNetV2 ``MatchedFrequency'' and `` Threshold 0.7'' are in Appendix~C.}). %
We show multi-label accuracies on two versions: ReaL~\cite{beyer2020are_we_done} and Shankar \etal~\cite{shanker2020machine_accuracy}. The metrics are identical: $\frac{1}{N}\sum_{n=1}^N 1(\arg\max f(x_n) \in y_n)$, where $1(\cdot)$ is the indicator function and $\arg\max f(x_n)$ is the top-1 prediction for a model $f$. The ground-truth multi-label for image $x_n$ is given as a set $y_n$. The difference between the metrics lies in the ground-truth multi-label annotation. 
We observe that \ours consistently achieves the best performance over all the metrics.
We obtain ${78.9\%}$ validation accuracy with ${+1.4}$ pp gain from the original labels, while the label smoothing and label cleaning boost only $+0.5$ pp and $+0.6$ pp, respectively.
On ImageNetV2, ReaL, and Shankar \etal metrics, \ours achieves ${80.5\%}$, ${85.0\%}$, and ${86.1\%}$ accuracies, where the gains are ${+1.5}$ pp, ${+1.4}$ pp, and ${+0.8}$ pp, respectively.
It is notable that only \ours achieves remarkable boosts on the multi-label benchmarks. Label smoothing and cleaning shows only marginal gains or even worse multi-label accuracies (\eg label cleaning results in a 0.1 pp worse result on Shankar \etal).
We confirm that \ours improves the performances of image classifiers and that it helps models truly learn to make better multi-label predictions.

\noindent\textbf{Results on various network architectures.}
We have trained various architectures with \ours to show that \ours is applicable to a wide range of networks with different training recipes. We consider ResNet-18, ResNet-101, EfficientNet-\{B0,B1,B2,B3\}~\cite{efficientnet}, and ReXNet~\cite{han2020rexnet}.
Training details to make the best performance out of EfficientNet models~\cite{efficientnet} are different from our base setting; we describe them in Appendix~D.2. %
We follow the original paper's training details for ReXNet~\cite{han2020rexnet}. 
Results are shown in Table~\ref{table:imagenet_various_arch}. \ours consistently enhances the performance of various network architectures. The 81.7\% accuracy of EfficientNet-B3 is further improved to 82.5\% with \ours.

\noindent\textbf{State-of-the-art performance.}
\ours is complementary to many other training tricks used for achieving the best model performances. For example, we combine a strong regularizer CutMix~\cite{yun2019cutmix} with \ours. CutMix mixes two training images via cut-and-paste manner and likewise mixes the labels. To use it with \ours, we perform CutMix on the randomly cropped images. The pooled labels are then mixed according to the CutMix algorithm. We set the hyper-parameter of CutMix $\alpha$ to $1.0$.
We show the results in Table~\ref{tab:towards-sota}.
\ours with CutMix achieves the state-of-the-art ImageNet top-1 accuracies of $\textbf{80.2\%}$ and $\textbf{81.6\%}$ for the ResNet-50 and ResNet-101 backbones. 
On top of this, we further consider using the extra training data based on the ImageNet-21K dataset~\cite{deng2009imagenet}: $14$M images with $21$K categories.
Unlike the previous work utilizing the ImageNet-21K~\cite{kolesnikov2019big} with their original single-class labels over 21K categories, we perform \ours on them to generate multi-labels over the $1$K classes.
We then sub-sample $4$M training data from the entire $14$M training images by balancing the top-1 class distributions, as done in~\cite{yalniz2019billion}.
Training with this extra data and CutMix on top of \ours boosts the accuracy of ResNet-50 to $\textbf{81.2\%}$.
In summary, \ours is a practical addition to existing training tricks that consistently improves the backbone performances.

\begin{table}
\vspace{-0.2cm}
\centering
\tabcolsep=0.12cm
\begin{tabular}{@{}lcc@{}}
\toprule
Model       & ImageNet top1 (\%) & \\ \midrule
ResNet-50 & 77.5  & \\
~~ + \oursb & 78.9 (+1.4) & \\
~~ + \oursb + CutMix & 80.2 (+2.7)& \\
~~ + \oursb + CutMix + Extra data & 81.2 (+3.7) & \\ \midrule
ResNet-101 & 78.1  & \\
~~ + \oursb & 80.7 (+2.6) & \\
~~ + \oursb + CutMix & 81.6 (+3.5) & \\
\bottomrule
\end{tabular}
\caption{\textbf{Towards the SOTA.} \ours with additional training tricks. ``Extra data'' refers to the ImageNet-21k dataset.}
\label{tab:towards-sota}
\vspace{-0.2cm}
\end{table}

\begin{figure}[t]
\centering
\includegraphics[width=\linewidth]{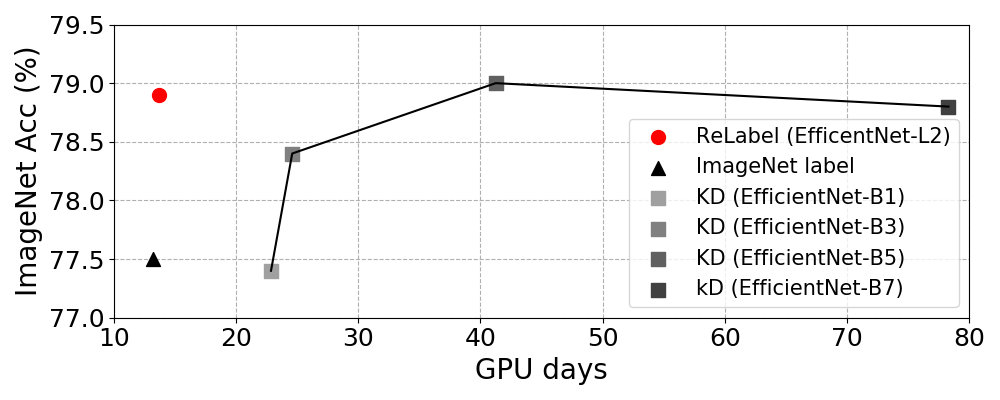}
\caption{
\textbf{Comparison against knowledge distillation.}
We plot ImageNet top-1 accuracies against the required training time for \ours and knowledge distillation (KD) approaches.
}
\label{fig:exp_kd_baseline}
\vspace{-0.2cm}
\end{figure}

\noindent\textbf{Comparison against knowledge distillation.}
We compare \ours against knowledge distillation (KD)~\cite{hinton2015distilling} in terms of the performance and training time costs. 
We train ResNet-50 with EfficientNet teachers: EfficientNet-\{B1,B3,B5,B7\}; we have not considered performing KD with EfficientNet-L2 as it would take 160 GPU days, beyond our computational capacity.
Training details for KD are in Appendix~D.3. %
Figure~\ref{fig:exp_kd_baseline} shows the results.
We plot the target model's performance versus the required number of GPU days.
KD with smaller teacher variants (EfficientNet-\{B1,B3\}) shows worse top-1 accuracies than \ours at higher training costs.
For larger teachers (EfficientNet-\{B5,B7\}), KD achieves comparable performances with \ours (\eg $79.0\%$ for KD with B7 and $78.8\%$ for \ours). 
However, they require $41$ and $78$ GPU days to train, compared to mere $13.6$ using \ours. \ours training is almost as fast as the original training.

\begin{table}[t]
  \centering
  \subfloat{
  \includegraphics[width=0.45\linewidth]{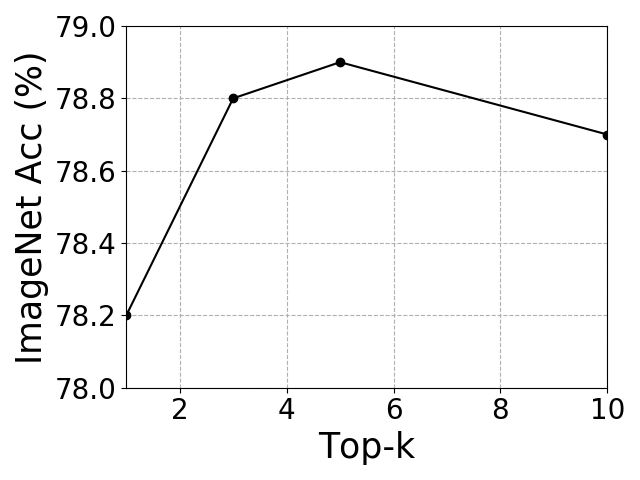}
  }%
  \qquad
  \subfloat{
        \tabcolsep=0.1cm
        \begin{tabular}{@{}lc@{}}
        \toprule
                     \multicolumn{2}{r}{ImageNet top1 (\%)}\\ \midrule
        Baseline & 77.5 \\
        \oursb        & 78.9  \\ \midrule
        16-bit label map  & 78.8 \\ 
         8-bit label map & 78.6 \\        
        \bottomrule
        \end{tabular}
  }
  \caption{\textbf{Storage versus performance.} How much performance do we lose by trying to cut storage space?}%
  \label{tab:exp_ablation}%
  \vspace{-0.2cm}
\end{table}

\noindent\textbf{Storage-performance trade off.}
We study the trade off between the storage space for the label maps and the model performance.
\ours only saves top-$k$ prediction maps for the interest of efficient storage, where the default $k$ value is 5. We explore $k \in \{1,3,5,10\}$. We also study the impact of quantization levels for label maps: $16$-bit and $8$-bit floating point, instead of the default $32$-bit floating point.
The results are in Table~\ref{tab:exp_ablation}.
\ours achieves a good performance-efficiency trade-off when $k=5$.
Quantizing label maps tend to yield only small performance drops ($-0.1$ pp to $-0.3$ pp).
When storage space is a crucial constraint, we advise users to adopt labels maps of coarser formats.

\noindent\textbf{Combination with original labels.}
When we combine \ours's annotation $L_\text{ours}\in[0,1]^C$ with the original label $L_\text{gt}\in\{0,1\}^C$ as $0.5L_\text{ours}+0.5L_\text{gt}$,
the performance degrades from 78.9\% to 78.3\% accuracy on ImageNet. 
They do not seem to make a good combination.

\noindent\textbf{Robustness.}
We evaluate the robustness of \ours-trained models against test-time perturbations. We consider adversarial and natural perturbations: FGSM~\cite{fgsm}, ImageNet-A~\cite{imagenet-a}, ImageNet-C~\cite{imagenet-c}, and background challenge (BGC)~\cite{bgc}. FGSM introduces one-step adversarial perturbations on images, while ImageNet-A samples consistent of common failure cases for modern image classifiers. ImageNet-C consists of 15 different types of natural perturbations.
BGC evaluates the robustness against backgrounds by selecting background images adversarially from the dataset.
Results are in Table~\ref{table:imagenet_robustness}.
We observe that \ours consistently improves the resilience of models on adversarial and natural perturbations. Especially, \ours shows remarkable improvements in the background robustness (+8.7\%) owing the localized supervision.
Furthermore, combining \ours with other training strategies, \eg, CutMix~\cite{yun2019cutmix} and extra training data, significantly boosts the performances in the all robustness benchmarks.

\newcommand{\snum}[1]{{\small #1}}

\begin{table}[t]
\centering
\resizebox{\columnwidth}{!} {
\tabcolsep=0.1cm
\begin{tabular}{@{}lcccc@{}}
\toprule
Models & FGSM & ImageNet-A & ImageNet-C & BCG \\ \midrule
ResNet-50 & 25.7  & 4.9  & 27.9 & 25.9 \\
+ {\small \oursb} & 31.3 \snum{(+5.6)} &  7.1 \snum{(+2.2)} & 28.1 \snum{(+0.2)} & 34.6 \snum{(+8.7)} \\
+ {\bf \small CutMix} & 42.4 \snum{(+16.7)} &  11.4 \snum{(+6.5)} & 47.5 \snum{(+19.6)} & 34.1 \snum{(+8.2)} \\
+ {\bf \small Extra data} & 45.0 \snum{(+19.3)} &  24.8 \snum{(+19.9)} & 54.2 \snum{(+26.3)} & 36.0 \snum{(+10.1)} \\
\bottomrule
\end{tabular}
}
\caption{\textbf{Robustness.} Impact of \ours on FGSM~\cite{fgsm}, ImageNet-A~\cite{imagenet-a}, ImageNet-C~\cite{imagenet-c}, and background challenge~\cite{bgc} benchmarks. All numbers are accuracies.}
\label{table:imagenet_robustness}
\vspace{-0.2cm}
\end{table}

\noindent\textbf{\ours examples on ImageNet.}
We present examples generated by \ours during ImageNet training in Appendix~E. %
As shown in the examples, ReLabel can generate location-specific multi-labels with more precise supervision than the original ImageNet labels.

\begin{table*}[t]
\tabcolsep=0.1cm
\centering
\begin{tabular}{lccccc}
\toprule
  & Food-101~\cite{food101} & Stanford Cars~\cite{stanford_cars} & DTD~\cite{cimpoi14describing} & FGVC Aircraft~\cite{fgvc_aircraft} & Oxford Pets~\cite{parkhi12a}\\ 
\midrule
ResNet-50 (Baseline)        & 87.98 & 92.64 & 75.43 & 85.09 & 93.92 \\
ResNet-50 (\oursb-trained)   & 88.12 & 92.73 & 75.74 & 88.89 & 94.28 \\ \bottomrule
\end{tabular}
\caption{\textbf{Fine-grained classification.} Performance on five tasks where the model starts either from weights regularly pre-trained on ImageNet or from weights pre-trained via \ours.}
\label{table:imagenet-finegrained}
\vspace{-0.2cm}
\end{table*}
\begin{table}[t]
\vspace{-0.2cm}
\small
\tabcolsep=0.08cm
\centering
\begin{tabular}{lcccc} 
\toprule
 & Faster-RCNN & & \multicolumn{2}{c}{Mask-RCNN}   \\ \cmidrule{2-2} \cmidrule{4-5}
 & bbox AP &  & bbox AP & mask AP \\ 
\midrule
{ResNet-50 (Baseline)}            & 37.7 & &  38.5 & 34.7 \\
{ResNet-50 (\oursb-trained)}      & 38.2 & & 39.1 & 35.2\\ 
\bottomrule
\end{tabular}
\caption{\textbf{Detection and instance segmentation.} Transfer learning performances for Faster-RCNN~\cite{fasterrcnn} and Mask-RCNN~\cite{he2017mask} on COCO dataset~\cite{lin2014microsoft}.}
\label{table:imagenet-detection}
\vspace{-0.2cm}
\end{table}

\subsection{Transfer Learning}
\label{sec:exp_transfer}

Apart from serving as the standard benchmark, ImageNet has contributed to the computer vision research and engineering with its suite of pre-trained models.
When the target task has only a small number of annotated data, transfer learning from the ImageNet pre-training usually helps~\cite{kornblith2019better}.
We examine here whether the \ours-induced improvements on the ImageNet performances transfer to various downstream tasks. 
We present the results of 5 fine-grained classification tasks and the object detection and instance segmentation tasks on COCO with models pre-trained on ImageNet with \ours.

\noindent\textbf{Fine-grained classification tasks.}
We evaluate \ours-pretrained ResNet-50 on five fine-grained classification tasks: Food-101~\cite{food101}, Stanford Cars~\cite{stanford_cars}, DTD~\cite{cimpoi14describing}, FGVC Aircraft~\cite{fgvc_aircraft}, and Oxford Pets~\cite{parkhi12a}. 
We use the standard data augmentation as in Section~\ref{sec:exp_imagenet_cls}. Models are fine-tuned with SGD for 5,000 iterations, 
following the convention for fine-tuning tasks~\cite{chatfield2014return}.
To find the best learning rate and weight decay values for each task, we perform a grid search per task and report the best performance.
Table~\ref{table:imagenet-finegrained} shows the results. 
Note that the \ours-trained model results in a consistent improvement over the vanilla pre-trained model. For example, on FGVC Aircraft, \ours pre-training improves the downstream task performance by $+3.8$ pp.

\noindent\textbf{Object detection and instance segmentation.}
We used Faster-RCNN~\cite{fasterrcnn} and Mask-RCNN~\cite{he2017mask} with feature pyramid network (FPN~\cite{lin2017feature}) as the base models for object detection and instance segmentation tasks, respectively.
The backbone networks of Faster-RCNN and Mask-RCNN are initialized with \ours-pretrained ResNet-50 model, and then fine-tuned on COCO dataset~\cite{lin2014microsoft} by the original training strategy~\cite{fasterrcnn,he2017mask} with the image size of $1200\times800$.
Table~\ref{table:imagenet-detection} shows the results. 
Pre-training with \ours improves the bbox AP of Faster-RCNN by $+0.5$ pp and the mask AP of Mask-RCNN by $+0.5$ pp. Pre-training a model with cleaner supervision like \ours leads to better feature representations and boosts the object detection and instance segmentation performances.

\subsection{Multi-label Classification}
\label{sec:exp_multilabel_cls}

\ours is designed to transform a single-label training set into a multi-label training set. 
Nonetheless, \ours and \ourframework also helps improving an originally multi-label training set by providing additional localized supervision signals, given that the random crop augmentation is a popular recipe for multi-label training as well~\cite{chen2019multi,wang2017multi,you2020cross}.
To see this effect, we experiment with the multi-label classification dataset COCO~\cite{lin2014microsoft} with multiple human-annotated labels per image. 
The baseline multi-label training uses multi-hot annotation $L\in\{0,1\}^C$ ($C=80$ for COCO).
Instead, we utilize the segmentation ground truth of COCO dataset as label maps $L\in\{0,1\}^{H\times W\times C}$ (\ie, an \emph{oracle} case of \ours). 
We also compare with {machine}-generated label maps $L\in\mathbb{R}^{H\times W\times C}$ from a state-of-the-art multi-label classifier~\cite{ben2020asymmetric} to see the effectiveness of the oracle label map.
We then train our multi-label classifiers with our \ourframework based on the label maps according to the random crop coordinates.
We conduct experiments on ResNet-50 and ResNet-101 networks with the input size $224\times224$ and $448\times448$, respectively, using the binary cross-entropy loss for all methods considered.
More training details are in Appendix~D.4. %
Table~\ref{table:multi-class} shows the results. 
We observe that applying \ours with machine-generated label maps results in $+3.7$ pp and $+2.4$ pp mAP gains and, with oracle label maps, \ours achieves more gain of {$\textbf{+4.2}$ pp} and {$\textbf{+4.3}$ pp} mAP gains on ResNet-50 and ResNet-101 networks, respectively.
In summary, the location-wise supervision from \ours helps the multi-label classification training.

\begin{table}
\vspace{-0.2cm}
\centering
\tabcolsep=0.12cm
\begin{tabular}{@{}lcc@{}}
\toprule
        & & COCO (mAP) \\
        \midrule
ResNet-50 & & 69.0 \\
ResNet-50 + \oursb (machine) &  &  72.7 \\
ResNet-50 + \oursb (oracle) &  &  73.2 \\ 
        \midrule
ResNet-101 & & 76.6 \\
ResNet-101 + \oursb (machine) &  &  79.0 \\
ResNet-101 + \oursb (oracle) &  &  80.9 \\ 
\bottomrule
\end{tabular}
\caption{\textbf{Originally multi-label tasks.} Results of \ours on COCO multi-class classification task~\cite{lin2014microsoft}.}
\label{table:multi-class}
\vspace{-0.2cm}
\end{table}

\section{Conclusion}
We have proposed a re-labeling strategy, \oursb, for the 1.28 million training images on ImageNet. \ours transforms the single-class labels assigned once per image into multi-class labels assigned for every region in an image, based on a machine annotator. The machine annotator is a strong classifier trained on a large extra source of visual data. We also proposed a novel scheme for training a classifier with the localized multi-class labels (\ourframeworkb). We experimentally verified significant performance gains induced by our labels and the corresponding training technique. \ours results in a consistent gain across tasks, including the ImageNet benchmarks, transfer-learning tasks, and multi-label classification tasks. We will open-source the localized multi-labels from \ours and the corresponding pre-trained models.

\vspace{0.1cm}
\noindent\textbf{Acknowledgement}
{
\small
We thank NAVER AI Lab members for valuable discussion and advice.
NAVER Smart Machine Learning (NSML)~\cite{nsml} has been used for experiments.
}

{\small
\bibliographystyle{ieee_fullname}
\bibliography{egbib}
}

\clearpage
\appendix
\onecolumn
\renewcommand{\thefigure}{A\arabic{figure}}
\setcounter{figure}{0}
\renewcommand{\thetable}{A\arabic{table}}
\setcounter{table}{0}
\renewcommand{\thealgorithm}{A\arabic{algorithm}}
\setcounter{algorithm}{0}

\section*{Appendix}

\section{\ours Algorithm}
\label{supp:algorithm}

\newcommand\graycomm[1]{{\textcolor{airforceblue}{#1}}}
\begin{algorithm*}[h]
  \caption{\ours Pseudo-code }
  \begin{algorithmic}[1]
    \For {each training iteration}
    \State \graycomm{\# Load image data and label maps (assume the minibatch size is $1$ for simplicity)}
    \State input, label\_map = \texttt{get\_minibatch}(dataset) 
    \State \graycomm{\# Random crop augmentation}
    \State [$c_x$,$c_y$,$c_w$,$c_h$] = \texttt{get\_crop\_region}(\texttt{size}(input))
    \State input = \texttt{random\_crop}(input, [$c_x$,$c_y$,$c_w$,$c_h$])
    \State input = \texttt{resize}(input, [$224,224$])
    \State \graycomm{\# \ourframework process}
    \State target = \texttt{RoIAlign}(label\_map, coords=[$c_x$,$c_y$,$c_w$,$c_h$], output\_size=($1,1$))
    \State target = \texttt{softmax}(target)
    \State \graycomm{\# Update model }
    \State output = \texttt{model\_forward}(input)
    \State loss = \texttt{cross\_entropy\_loss}(output, target)
    \State \texttt{model\_update}(loss)
    \EndFor
  \end{algorithmic}
  \label{alg:supp_algorithm}
\end{algorithm*}

We present the pseudo-codes of \ours in Algorithm~\ref{alg:supp_algorithm}.
We assume the minibatch size is $1$ for simplicity. 
First, an input image and its saved label map are loaded from the dataset. 
Then the random crop augmentation is conducted on the input image. 
We then perform \texttt{RoIAlign} on the label map with the random crop coordinates [$c_x$,$c_y$,$c_w$,$c_h$].
Finally \texttt{softmax} function is conducted on the pooled label map to get a multi-label ground-truth in $[0, 1]^C$.
The multi-label ground-truth is used for updating the model with the standard cross-entropy loss. 

\begin{figure*}[h]
    \centering
    \includegraphics[width=\linewidth]{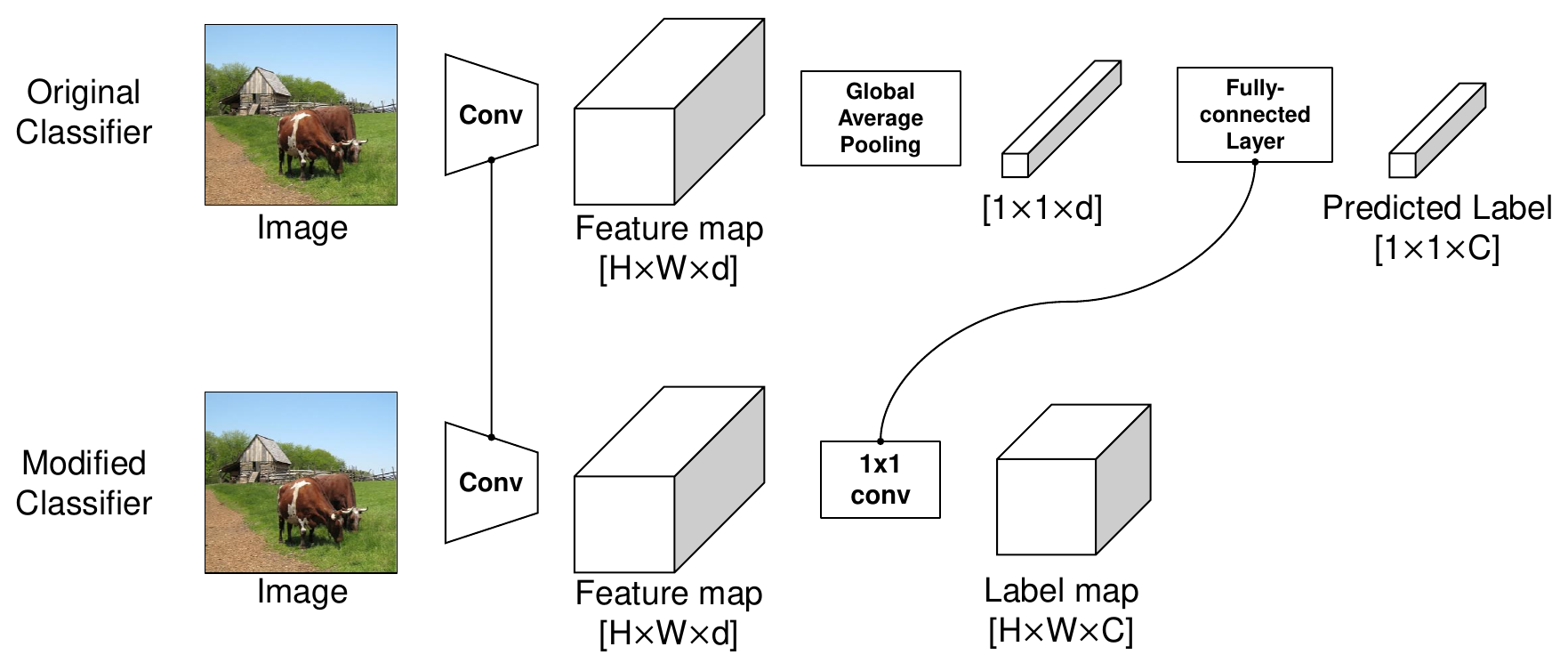}
    \caption{\textbf{Obtaining a label map.}
    The original classifier (upper) takes an input image and generates a predicted label $L_\text{org}\in \mathbb{R}^{1\times1\times C}$. 
    On the other hand, the modified classifier (lower) outputs a \textit{label map} $L_\text{ours}\in \mathbb{R}^{H\times W \times C}$ by removing the \textit{global average pooling} layer. 
    Note that the ``Fully-connected Layer'' ($\mathbf{W}_\text{fc} \in \mathbb{R}^{d\times C}$) of the original classifier and ``$1\times1$ conv'' ($\mathbf{W}_\text{1x1 conv} \in \mathbb{R}^{1\times1\times d\times C}$) of the modified classifier are identical.
    }
    \label{fig:supp_label_map}
\end{figure*}

\begin{figure*}[h]
    \centering
    \includegraphics[width=1.0\linewidth]{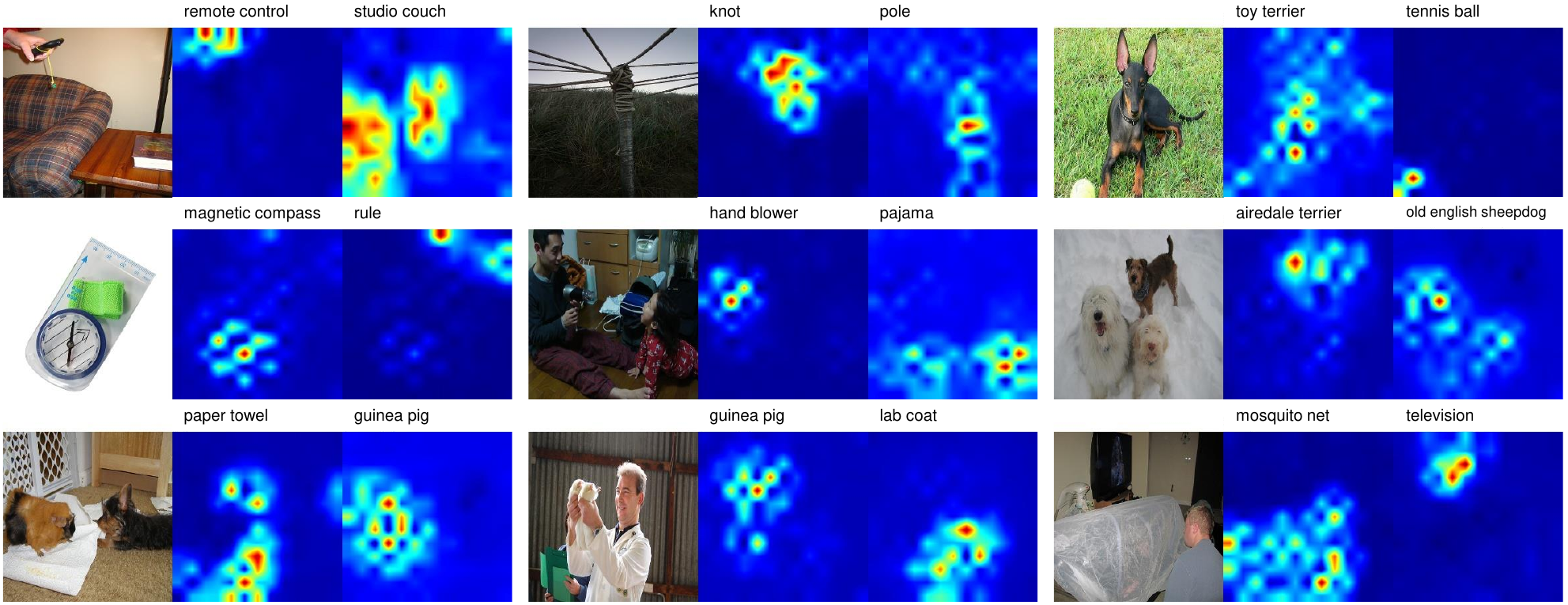}
    \caption{\textbf{Label map examples.}
    Each example presents the input image (left), label map of top-1 class (middle), label map of top-2 class (right). 
    }
    \label{fig:supp_label_map_example}
\end{figure*}

\section{Re-labeling ImageNet: Detailed Procedure and Examples}
\label{supp:label_maps}

We show the detailed process of obtaining a label map in Figure~\ref{fig:supp_label_map}.
The original classifier takes an input image, computes the feature map ($\mathbb{R}^{H\times W \times d}$), conducts \textit{global average pooling} ($\mathbb{R}^{1\times1\times d}$), and generates the predicted label $L_\text{org}\in \mathbb{R}^{1\times1\times C}$ with the fully-connected layer ($\mathbf{W}_\text{fc} \in \mathbb{R}^{d\times C}$). 
On the other hand, the modified classifier do not have \textit{global average pooling} layer, and outputs a \textit{label map} $L_\text{ours}\in \mathbb{R}^{H\times W \times C}$ from the feature map ($\mathbb{R}^{H\times W \times d}$). 
Note that the fully-connected layer ($\mathbf{W}_\text{fc} \in \mathbb{R}^{d\times C}$) of the original classifier and $1\times1$ conv ($\mathbf{W}_\text{1x1 conv} \in \mathbb{R}^{1\times1\times d\times C}$) of the modified classifier are identical.

We utilize EfficientNet-L2~\cite{xie2020noisy_student} as our machine annotator classifier whose input size is $475 \times 475$. 
For all training images, we resize them into $475 \times 475$ without cropping and generate label maps by feed-forwarding them. 
The spatial size of label map $(W,H)$ is $(15,15)$, number of channel $d$ is $5504$, and the number of classes $C$ is $1000$.

We present several label map examples in Figure~\ref{fig:supp_label_map_example}.
From a label map $L\in \mathbb{R}^{H\times W \times C}$, we only show two heatmaps for the classifier's top-2 categories. The heatmap is $L[c_i,:,:] \in \mathbb{R}^{H\times W}$ where $c_i$ is one of the top-2 categories. 
As shown in the examples, the top-1 and top-2 heatmaps are disjointly located at each object's position.

\section{Results on ImageNetV2}
\label{supp:imagenet_v2}

\begin{table*}
\centering
\tabcolsep=0.1cm
\begin{tabular}{@{}llcccc@{}}
\toprule
& & ImageNet &  ImageNetV2 & ImageNetV2 &  ImageNetV2\\
Network\hspace{2em} & Supervision & & {\small (Top-Images)} &  {\small (Matched Frequency)} & {\small (Threshold 0.7)} \\
            
\midrule
ResNet-50 & Original  & 77.5 & 79.0 & 65.2 & 74.3 \\
ResNet-50 & Label smoothing ($\epsilon$=0.1)~\cite{szegedy2016rethinking_labelsm} & 78.0 & 79.5 & 66.0 & 74.6 \\
ResNet-50 & Label cleaning~\cite{beyer2020are_we_done}  & 78.1 & 79.1 & 64.9 & 73.9 \\ 
ResNet-50 & \oursb   & \textbf{78.9} & \textbf{80.5} & \textbf{67.3} & \textbf{76.0}  \\ 
\bottomrule
\end{tabular}
\caption{\textbf{ImageNetV2 results.} 
We report performances on ImageNetV2~\cite{recht2019imagenetv2} metrics.
}
\label{table:supp_imagenetv2}
\end{table*}

We present full ImageNetV2~\cite{recht2019imagenetv2} results in Table~\ref{table:supp_imagenetv2}.
Three metrics ``Top-Images'', ``Matched Frequency'', and ``Threshold 0.7'' are reported with two baselines Label smoothing~\cite{szegedy2016rethinking_labelsm} and Label cleaning~\cite{beyer2020are_we_done}. 
\ours obtained $80.5$, $67.3$, and $76.0$ accuracies on ImageNetV2 ``Top-Images'', ``Matched Frequency'', and ``Threshold 0.7'', where the gains are $+1.5$, $+2.1$, and $+1.7$ pp against the vanilla ResNet-50, respectively.  

\section{Implementation details}
\label{supp:impl_details}
We present the implementation details in this section. 

\subsection{Training Hyper-parameters}
\label{supp:impl_details_hyperparameters}
In most experiments, we have trained the models with SGD optimizer with learning rate $0.1$ and weight decay $0.0001$.
For further improved performance, we utilized AdamP~\cite{heo2020adamp} optimizer with learning rate $0.002$, and weight decay $0.01$ when applying additional tricks such as CutMix regularizer or extra training data (ImageNet-21K). 

\subsection{EfficientNet on ImageNet}
\label{supp:impl_details_efficientnet}
We utilize an open-source pytorch codebase~\cite{timm} to train EfficientNet variants on ImageNet.
We utilize AdamP~\cite{heo2020adamp} optimizer and set training epochs $400$, minibatch size $512$, learning rate $0.002$, and weight decay $0.01$ with four NVIDIA V100 GPUs.
Dropout and drop path~\cite{stochasticdepth} regularizers are used with dropout rate $0.2$ and drop path rate $0.2$, respectively.
We also utilize Random erasing~\cite{zhong2017randomerase}, RandAugment~\cite{cubuk2020randaugment}, and Mixup~\cite{zhang2017mixup} augmentations as suggested in \cite{timm}.
All training settings are samely used for both vanilla training and \ours training of EfficientNet variants.

\subsection{Knowledge Distillation}
\label{supp:impl_details_kd}
Training with knowledge distillation is also conducted on the pytorch codebase~\cite{timm}.
For teacher network, we use official EfficientNet (B1-B7)~\cite{efficientnet} weights trained with noisy student~\cite{xie2020noisy_student} techniques.
We utilize outputs of networks after soft-max layer and the cross-entropy loss between teacher and students is only used for the distillation loss~\cite{hinton2015distilling}.
The temperature and cross-entropy with ground truth were not used.
Since the EfficientNet teachers are trained with large-size images ($240\times240-600\times600$), we put the large-size image for teacher network and resize it to $224\times224$ for inputs of student network.
We adopt SGD with Nesterov momentum for the optimizer and the standard setting~\cite{resnet} with long epochs: learning rate $0.1$, weight decay $10^{-4}$, batch size $256$, training epochs $300$ and cosine learning rate schedule with four NVIDIA V100 GPUs.

\subsection{COCO Multi-label Classification}
\label{supp:impl_details_cocomulti}

As in recent multi-label classification works~\cite{chen2019multi,wang2017multi,you2020cross,ben2020asymmetric}, the classifier model is initialized with ImageNet-pretrained model and fine-tuned on COCO multi-label dataset~\cite{lin2014microsoft}.
We utilize the official pytorch ImageNet-pretrained model using {torchvision} toolbox\footnote{https://github.com/pytorch/vision}.
The weight of the final fully-connected layer is modified from $\mathbb{R}^{d\times 1000}$ to $\mathbb{R}^{d\times 80}$ to fit the number of classes for COCO dataset and the weight matrix is randomly initialized.
For fine-tuning, we utilize AdamP~\cite{heo2020adamp} optimizer and cosine learning rate schedule with initial learning rate $0.0002$ and weight decay $0.01$.
We set the minibatch size to $128$.
The input resolution is $224 \times 224$ for ResNet-50 and $448\times448$ for for ResNet-101.
To obtain machine-generated label maps, we utilize a pre-trained TResNet-XL model~\cite{ben2020asymmetric} whose input size is $640 \times 640$ and mAP is $88.4\%$.
\section{\ours Examples on ImageNet}
\label{supp:sec:relabel_examples}

\begin{figure}
    \centering
    \includegraphics[width=\linewidth]{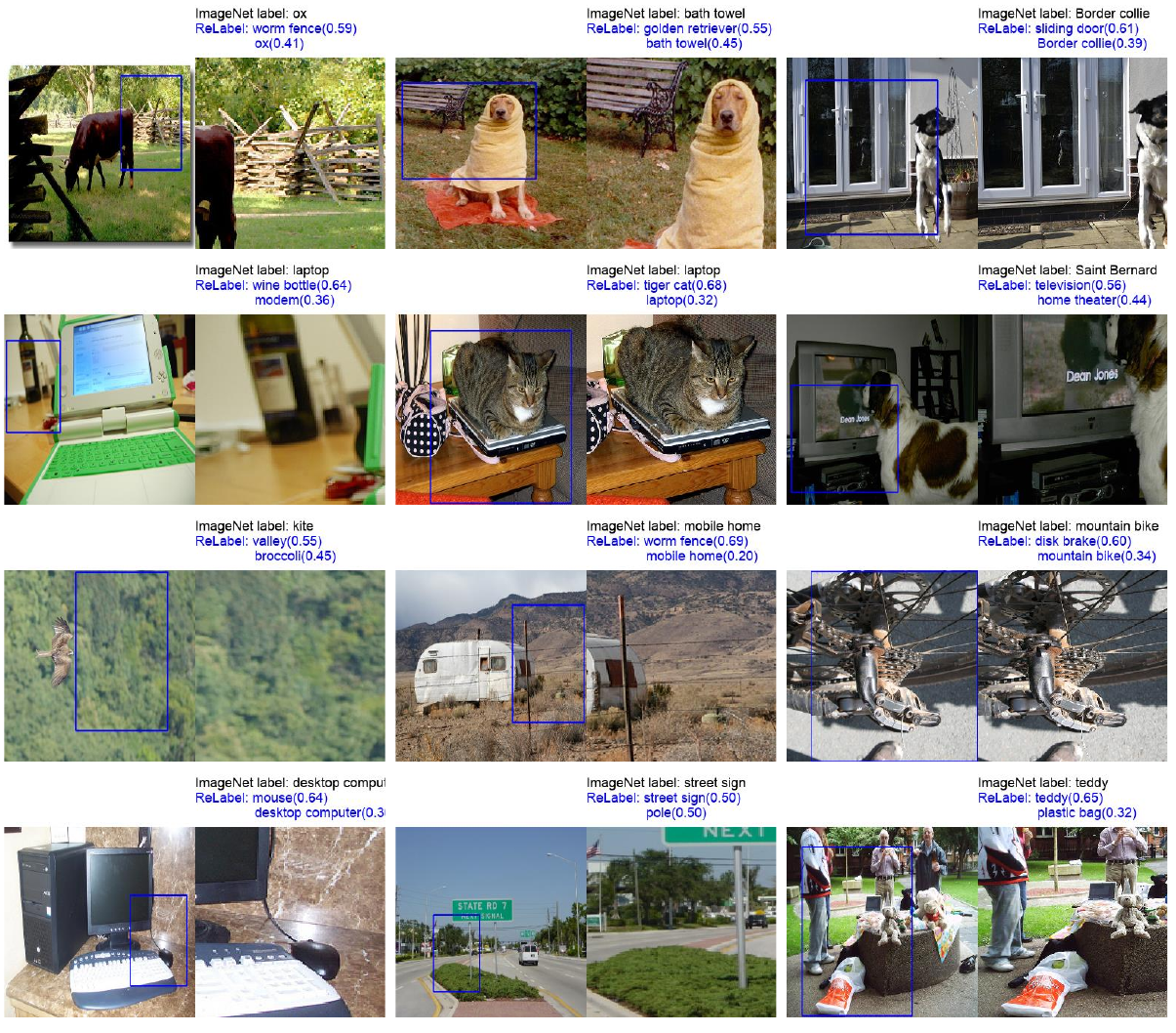}
    \caption{\textbf{\ours examples during ImageNet training.}
    We present selected examples generated by \ours during ImageNet training. 
    For each example, the left image is the full training image and the right image is the random cropped patch. The random crop coordinates are denoted by blue bounding boxes. The original ImageNet label and \ours are also presented. 
    }
    \label{fig:supp_relabel_samples}
\end{figure}

We present \ours exsamples on ImageNet training data in Figure~\ref{fig:supp_relabel_samples}.
We show the full training images (left) and the random cropped patches (right). 
The random crop coordinates are denoted by blue bounding boxes. 
We also present the original ImageNet label and the new multi-labels by \ours. 
As shown in the examples, \ours can generate location-specific multi-labels with more precise supervision than the original ImageNet label.

\end{document}